# Three-dimensional virtual refocusing of fluorescence microscopy images using deep learning

**Authors:**

Yichen Wu[1,2,3,†], Yair Rivenson[1,2,3,†], Hongda Wang[1,2,3], Yilin Luo[1,2,3], Eyal Ben-David[4], Laurent A. Bentolila[3,5], Christian Pritz[6], Aydogan Ozcan[1,2,3,7,*]

[1] Electrical and Computer Engineering Department, University of California, Los Angeles, California 90095, USA

[2] Bioengineering Department, University of California, Los Angeles, California 90095, USA

[3] California Nano Systems Institute (CNSI), University of California, Los Angeles, California 90095, USA

[4] Department of Human Genetics, David Geffen School of Medicine, University of California, Los Angeles, California 90095, USA

[5] Department of Chemistry and Biochemistry, University of California, Los Angeles, CA, 90095, USA

[6] Department of Genetics, Hebrew University of Jerusalem, Edmond J. Safra Campus, Givat Ram, Jerusalem, 91904, Israel

[7] Department of Surgery, David Geffen School of Medicine, University of California, Los Angeles, California 90095, USA

[†]Equal contribution authors

[*]Corresponding author, email: ozcan@ucla.edu

**Abstract:**

Three-dimensional (3D) fluorescence microscopy in general requires axial scanning to capture images of a sample at different planes. Here we demonstrate that a deep convolutional neural network can be trained to virtually refocus a 2D fluorescence image onto user-defined 3D surfaces within the sample volume. With this data-driven computational microscopy framework, we imaged the neuron activity of a *Caenorhabditis elegans* worm in 3D using a time-sequence of fluorescence images acquired at a single focal plane, digitally increasing the depth-of-field of the microscope by 20-fold without any axial scanning, additional hardware, or a trade-off of imaging resolution or speed. Furthermore, we demonstrate that this learning-based approach can correct for sample drift, tilt, and other image aberrations, all digitally performed after the acquisition of a single fluorescence image. This unique framework also cross-connects different imaging modalities to each other, enabling 3D refocusing of a single wide-field fluorescence image to match confocal microscopy images acquired at different sample planes. This deep learning-based 3D image refocusing method might be transformative for imaging and tracking of 3D biological samples, especially over extended periods of time, mitigating photo-toxicity, sample drift, aberration and defocusing related challenges associated with standard 3D fluorescence microscopy techniques.

**Text:**

Three-dimensional (3D) fluorescence microscopic imaging is essential for biomedical and physical sciences as well as engineering, covering various applications [1–7]. Despite its broad importance, high-throughput acquisition of fluorescence image data for a 3D sample remains a challenge in microscopy research. 3D fluorescence information is usually acquired through scanning across the sample volume, where several 2D fluorescence images/measurements are obtained, one for each focal plane or point in 3D , which forms the basis of e.g., confocal [8], two-photon [5], light-sheet [7,9,10] or various super-resolution [2,4,11–14] microscopy techniques. However, because scanning is used, the image acquisition speed and the throughput of the system for volumetric samples are limited to a fraction of the frame-rate of the camera/detector, even with optimized scanning strategies [8] or point-spread function (PSF) engineering [7,15]. Moreover, because the images at different sample planes/points are not acquired simultaneously, the temporal variations of the sample fluorescence can inevitably cause image artifacts. Another concern is the photo-toxicity of illumination and photo-bleaching of fluorescence since parts of the sample can be repeatedly excited during the scanning process.

To overcome some of these challenges, non-scanning 3D fluorescence microscopy methods have also been developed, so that the entire 3D volume of the sample can be imaged at the same speed as the detector framerate. One of these methods is fluorescence light-field microscopy [6,16–19]. This system typically uses an additional micro-lens array to encode the 2D angular information as well as the 2D spatial information of the sample light rays into image sensor pixels; then a 3D focal stack can be digitally reconstructed from this recorded 4D light-field. However, using a micro-lens array reduces the spatial sampling rate which results in a sacrifice of both the lateral and axial resolution of the microscope. Although the image resolution can be improved by 3D deconvolution [6] or compressive sensing [17] techniques, the success of these methods depends on various assumptions regarding the sample and the forward model of the image formation process. Furthermore, these computational approaches are relatively time-consuming as they involve an iterative hyperparameter tuning as part of the image reconstruction process. A related method termed multi-focal microscopy has also been developed to map the depth information of the sample onto different parallel locations within a single image [20,21]. However, the improved 3D imaging speed of this method also comes at the cost of reduced imaging resolution or field-of-view (FOV) and can only infer an experimentally pre-defined (fixed) set of focal planes within the sample volume. As another alternative, the fluorescence signal can also be optically correlated to form a Fresnel correlation hologram, encoding the 3D sample information in interference patterns [22–24]. To retrieve the missing phase information, this computational approach requires multiple images to be captured for volumetric imaging of a sample. Quite importantly, all these methods summarized above, and many others, require the addition of customized optical components and hardware into a standard fluorescence microscope, potentially needing extensive alignment and calibration procedures, which not only increase the cost and complexity of the optical set-up, but also cause potential aberrations and reduced photon-efficiency for the fluorescence signal.

Here we introduce a digital image refocusing framework in fluorescence microscopy by training a deep neural network using microscopic image data, enabling 3D imaging of fluorescent samples using a single 2D wide-field image, without the need for any mechanical scanning, additional hardware or parameter estimation. This data-driven fluorescence image refocusing framework does not need a physical model of the imaging system, and rapidly refocuses a 2D fluorescence image onto user-defined 3D surfaces. In addition to rapid 3D imaging of a fluorescent sample volume, it can also be used to digitally correct for various aberrations due to the sample and/or the optical system. We term this deep learning-based approach *Deep-Z*, and use it to computationally refocus a single 2D wide-field fluorescence image onto 3D surfaces within the sample volume, without sacrificing the imaging speed, spatial resolution, field-of-view, or throughput of a standard fluorescence microscope.

In *Deep-Z* framework, an input 2D fluorescence image (to be digitally refocused onto a 3D surface within the sample volume) is first appended with a user-defined digital propagation matrix (DPM) that represents, pixel-by-pixel, the axial distance of the target surface from the plane of the input image (Fig. 1). *Deep-Z* is trained using a conditional generative adversarial neural network (GAN) [25,26] using accurately matched pairs of (1) various fluorescence images axially-focused at different depths and appended with different DPMs, and (2) the corresponding fluorescence images (i.e., the ground truth labels) captured at the correct/target focus plane defined by the corresponding DPM. Through this training process that only uses experimental image data without any assumptions or physical models, the generator network of GAN *learns to interpret the values of each DPM pixel as an axial refocusing distance*, and outputs an equivalent fluorescence image that is digitally refocused within the sample volume to the 3D surface defined by the user, where some parts of the sample are focused, while some other parts get out-of-focus, according to their true axial positions with respect to the target surface.

To demonstrate the success of this unique fluorescence digital refocusing framework, we imaged *Caenorhabditis elegans* (*C. elegans*) neurons using a standard wide-field fluorescence microscope with a 20×/0.75 numerical aperture (NA) objective lens, and extended the native depth-of-field (DOF) of this objective (~1 µm) by ~20-fold, where a single 2D fluorescence image was axially refocused using *Deep-Z* to $\Delta z = \pm 10$ µm with respect to its focus plane, providing a very good match to the fluorescence images acquired by mechanically scanning the sample within the same axial range. Similar results were also obtained using a higher NA objective lens (40×/1.3 NA). Using this deep learning-based fluorescence image refocusing technique, we further demonstrated 3D tracking of the neuron activity of a *C. elegans* worm over an extended DOF of ±10 μm using a time-sequence of fluorescence images acquired at a *single* focal plane.

Furthermore, to highlight some of the additional degrees-of-freedom enabled by *Deep-Z*, we used spatially *non-uniform* DPMs to refocus a 2D input fluorescence image onto user-defined 3D surfaces to computationally correct for aberrations such as sample drift, tilt and spherical aberrations, all performed *after* the fluorescence image acquisition and without any modifications to the optical hardware of a standard wide-field fluorescence microscope.

Another important feature of this *Deep-Z* framework is that it permits *cross-modality digital refocusing* of fluorescence images, where the GAN is trained with gold standard label images obtained by a *different* fluorescence microscopy modality to teach the generator network to refocus an input image onto another plane within the sample volume, but this time to match the image of the same plane that is acquired by a different fluorescence imaging modality compared to the input image. We term this related framework *Deep-Z+*. To demonstrate the proof-of-concept of this unique capability, we trained *Deep-Z+* with input and label images that were acquired with a wide-field fluorescence microscope and a *confocal* microscope, respectively, to blindly generate at the output of this cross-modality *Deep-Z+*, digitally refocused images of an input wide-field fluorescence image that match confocal microscopy images of the same sample sections.

After its training, *Deep-Z* remains fixed, while the appended DPM provides a "depth tuning knob" for the user to refocus a single 2D fluorescence image onto 3D surfaces and output the desired digitally-refocused fluorescence image in a rapid non-iterative fashion. In addition to fluorescence microscopy, *Deep-Z* framework might potentially be applied to other incoherent imaging modalities, and in fact it bridges the gap between coherent and incoherent microscopes by enabling 3D digital refocusing of a sample volume using a single 2D incoherent image. *Deep-Z* is further unique in that it enables a computational framework for rapid transformation of a 3D surface onto another 3D surface within the fluorescent sample volume using a single forward-pass operation of the generator network.

**Digital refocusing of fluorescence images using *Deep-Z***

*Deep-Z* enables a single intensity-only wide-field fluorescence image to be digitally refocused to a user-defined surface within the axial range of its training (see the Methods for training details). Fig. 1a demonstrates this concept by digitally propagating a single fluorescence image of a 300 nm fluorescent bead (excitation/emission: 538 nm/584 nm) to multiple user defined planes. The native DOF of the input fluorescence image, defined by the NA of the objective lens (20×/0.75 NA), is ~ 1 µm; using *Deep-Z*, we were able to digitally refocus the image of this fluorescent bead over an axial range of ~ ±10 µm, matching the mechanically-scanned corresponding images of the same region of interest (ROI), which form the ground truth. Note that the PSF in Fig. 1a is asymmetric in the axial direction, which provides *directional* cues to the neural network regarding the digital propagation of an input image by *Deep-Z*. Unlike a symmetric Gaussian beam[27], such PSF asymmetry along the axial direction is ubiquitous in fluorescence microscopy systems[28]. In addition to digitally refocusing an input fluorescence image, *Deep-Z* also provides improved signal-to-noise ratio (SNR) at its output in comparison to a fluorescence image of the same object measured at the corresponding depth (see Supplementary Fig. S1); at the heart of this SNR increase compared to a mechanically-scanned ground truth is the ability of the neural network to reject various sources of image noise that were not generalized during its training phase. To further quantify *Deep-Z* output performance we used PSF analysis; Figs. 1b,c illustrate the histograms of both the *lateral* and the *axial* full-width-half-maximum (FWHM) values of 461 individual/isolated nano-beads distributed over ~ 500 × 500 µm$^2$. The statistics of these histograms very well agree with each other (Fig. 1b,c), confirming the match between *Deep-Z* output images calculated from a single fluorescence image (*N*=1 measured image) and the corresponding axially-scanned ground truth images (*N*=41 measured images). This quantitative match highlights the fact that *Deep-Z* indirectly learned, through image data, the 3D refocusing of fluorescence light. However, this learned capability is limited to be within the axial range determined by the training dataset (e.g., ±10 µm in this work), and fails outside of this training range (see Supplementary Fig. S2 for quantification of this phenomenon).

Next, we tested the *Deep-Z* framework by imaging the neurons of a *C. elegans* nematode expressing pan-neuronal tagRFP [29]. Fig. 2 demonstrates our blind testing results for *Deep-Z* based refocusing of different parts of a *C. elegans* worm from a *single* wide-field fluorescence input image. Using *Deep-Z*, non-distinguishable fluorescent neurons in the input image were brought into focus at different depths, while some other in-focus neurons at the input image got out-of-focus and smeared into the background, according to their true axial positions in 3D; see the cross-sectional comparisons to the ground truth mechanical scans provided in Fig. 2 (also see Supplementary Fig. S3 for image difference analysis). For optimal performance, this *Deep-Z* was specifically trained using *C. elegans* samples, to accurately learn the 3D PSF information together with the refractive properties of the nematode body and the surrounding medium. Using *Deep-Z*, we also generated (from a single 2D fluorescence image of a *C. elegans* worm) a virtual 3D stack (Video S1) and 3D visualization (Video S2) of the sample, over an axial range of ~ ±10 µm. Similar results were also obtained for a *C. elegans* imaged under a 40×/1.3NA objective lens, where *Deep-Z* successfully refocused the input image over an axial range of ~ ±4 µm; see Supplementary Fig. S4 for details.

Since *Deep-Z* can digitally reconstruct the image of an arbitrary plane within a 3D sample using a single 2D fluorescence image, without sacrificing the inherent resolution, frame-rate or photon-efficiency of the imaging system, it is especially useful for imaging dynamic biological samples. To demonstrate this capability, we captured a video of four moving *C. elegans* worms, where each frame of this fluorescence video was digitally refocused to various depths using *Deep-Z*. This enabled us to create simultaneously running videos of the same sample volume, each one being focused at a different depth (see Video S3). This unique capability enabled by *Deep-Z* not only eliminates the need for mechanical axial scanning and related optical hardware, but also significantly reduces phototoxicity or photobleaching within the sample to enable longitudinal experiments. Furthermore, each one of these virtually created videos are by definition temporally synchronized to each other (i.e., the frames at different depths have identical time stamps), which is not possible with a scanning-based 3D imaging system due to the unavoidable time delay between successive measurements of different parts of the sample volume. In addition to 3D imaging of the neurons of a nematode, *Deep-Z* also works well to digitally refocus the images of fluorescent samples that are spatially denser such as the mitochondria and F-actin structures within bovine pulmonary artery endothelial cells (BPAEC); see Supplementary Fig. S5 for examples of these results.

In our results reported so far, the blindly tested samples were inferred with a *Deep-Z* network that was trained using the same type of sample and the same microscopy system. In the Supplementary Information, under the sub-sections '*Robustness of Deep-Z to changes in samples and imaging systems*' as well as '*Deep-Z virtual refocusing capability at lower image exposure*', we evaluated the performance of *Deep-Z* under different scenarios, where a change in the test data distribution is introduced in comparison to the training image set, such as e.g., (1) a different type of sample is imaged, (2) a different microscopy system is used for imaging, and (3) a different illumination power or SNR is used; also refer to Supplementary Figs. S10-12. Our results and related analysis reveal the robustness of *Deep-Z* to some of these changes; however, as a general recommendation to achieve the best performance with *Deep-Z*, the neural network should be trained (from scratch or through transfer learning, which significantly expedites the training process, as illustrated in Supplementary Figs. S11-12) using training images obtained with the same microscope system and the same types of samples, as expected to be used at the testing phase.

**Sample drift-induced defocus compensation using *Deep-Z***

*Deep-Z* also enables the correction for sample drift induced defocus after the image is captured. To demonstrate this, Video S4 shows a moving *C. elegans* worm recorded by a wide-field fluorescence microscope with a 20×/0.8NA objective lens (DOF ~ 1 µm). The worm was defocused ~ 2 – 10 µm from the recording plane. Using *Deep-Z*, we can digitally refocus each frame of the input video to different planes up to 10 µm, correcting this sample drift induced defocus (Video S4). Such a sample drift is conventionally compensated by actively monitoring the image focus and correcting for it during the measurement, e.g., by using an additional microscope[8]. *Deep-Z*, on the other hand, provides the possibility to compensate sample drift in already-captured 2D fluorescence images.

## 3D functional imaging of *C. elegans* using *Deep-Z*

An important application of 3D fluorescence imaging is neuron activity tracking; for example genetically modified animals that express different fluorescence proteins are routinely imaged using a fluorescence microscope to reveal their neuron activity. To highlight the utility of *Deep-Z* for tracking the activity of neurons in 3D, we recorded the fluorescence video of a *C. elegans* worm at a single focal plane ($z = 0$ µm) at ~3.6 Hz for ~35 sec, using a 20×/0.8NA objective lens with two fluorescence channels: FITC for neuron activity and Texas Red for neuron locations. The input video frames were registered with respect to each other to correct for the slight body motion of the worm between the consecutive frames (see the Methods for details). Then, each frame at each channel of the acquired video were digitally refocused using *Deep-Z* to a series of axial planes from -10 µm to 10 µm with 0.5 µm step size, generating a virtual 3D fluorescence stack for each acquired frame. Video S5 shows a comparison of the recorded input video and a video of the maximum intensity projection (MIP) along *z* for these virtual stacks. As can be seen in this comparison, the neurons that are defocused in the input video can be clearly refocused on demand at the *Deep-Z* output for both of the fluorescence channels. This enables accurate spatio-temporal tracking of individual neuron activity in 3D from a temporal sequence of 2D fluorescence images, captured at single focal plane.

To quantify the neuron activity using *Deep-Z* output images, we segmented voxels of each individual neuron using the Texas Red channel (neuron locations), and tracked the change of the fluorescence intensity, i.e., $\Delta F(t) = F(t) - F_0$, in the FITC channel (neuron activity) inside each neuron segment over time, where $F(t)$ is the neuron fluorescence emission intensity and $F_0$ is its time average; see Methods for details. A total of 155 individual neurons in 3D were isolated using *Deep-Z* output images, as shown in Fig. 3b, where the color represents the depth (*z* location) of each neuron; also see Video S6. For comparison, we also report in Supplementary Fig. S14(b) the results of the same segmentation algorithm applied on just the input 2D image, where 99 neurons were identified, without any depth information; see the Supplementary Information for further analysis reported in the sub-section "*C. elegans neuron segmentation comparison*".

Fig. 3c plots the activities of the 70 most active neurons, which were grouped into clusters C1-C3 based on their calcium activity pattern similarities (see Methods for details). The activities of all of the 155 neurons inferred using *Deep-Z* are provided in Supplementary Fig. S6. Fig. 3c reports that cluster C3 calcium activities increased at $t = 14$ s, whereas the activities of cluster C2 decreased at a similar time point. These neurons very likely correspond to the motor neurons type A and B that promote backward and forward motion, respectively, which typically anti-correlate with each other[30]. Cluster C1 features two cells that were comparatively larger in size, located in the middle of the worm. These cells had three synchronized short spikes at t = 4, 17 and 32 sec. Their 3D positions and calcium activity pattern regularity suggest that they are either neuronal or muscle cells of the defecation system that initiates defecation in regular intervals in coordination with the locomotion system[31].

We emphasize that all this 3D tracked neuron activity was in fact embedded in the input 2D fluorescence image sequence acquired at a single focal plane within the sample, but could not be readily inferred from it. Through the *Deep-Z* framework and its 3D refocusing capability to user defined surfaces within the sample volume, the neuron locations and activities were accurately tracked using a 2D microscopic time sequence, without the need for mechanical scanning, additional hardware, or a trade-off of resolution or imaging speed.

Since *Deep-Z* generates temporally synchronized virtual image stacks through purely digital refocusing, it can be used to match the imaging speed to the limit of the camera framerate, by using e.g., the stream mode, which typically enables a short video of up to 100 frames per second. To highlight this opportunity, we used the stream mode of the camera of a Leica SP8 microscope (see the Methods section for details) and captured two videos at 100 fps for monitoring the neuron nuclei (under the Texas Red channel) and the neuron calcium activity (under the FITC channel) of a moving *C. elegans* over a period of 10 sec, and used *Deep-Z* to generate virtually refocused videos from these frames over an axial depth range of +/- 10 µm, as shown in Video S7 and Video S8, respectively.

### *Deep-Z* based aberration correction using spatially non-uniform DPMs

The results that we reported up to now used *uniform* DPMs in both the training phase and the blind testing in order to refocus an input fluorescence image to different planes within the sample volume. Here we emphasize that, even though *Deep-Z* was trained with uniform DPMs, in the testing phase one can also use spatially *non-uniform* entries as part of a DPM to refocus an input fluorescence image onto user-defined 3D surfaces. This capability enables digital refocusing of the fluorescence image of a 3D surface onto another 3D surface, defined by the pixel mapping of the corresponding DPM.

Such a unique capability can be useful, among many applications, for simultaneous auto-focusing of different parts of a fluorescence image after the image capture, measurement or assessment of the aberrations introduced by the optical system (and/or the sample) as well as for correction of such aberrations by applying a desired non-uniform DPM. To exemplify this additional degree-of-freedom enabled by *Deep-Z*, Fig. 4 demonstrates the correction of the planar tilting and cylindrical curvature of two different samples, after the acquisition of a single 2D fluorescence image per object. Fig. 4a illustrates the first measurement, where the plane of a fluorescent nano-bead sample was tilted by 1.5° with respect to the focal plane of the objective lens (see Methods for details). As a result, the left and right sides of the acquired raw fluorescence image (Fig. 4c) were blurred and the corresponding lateral FWHM values for these nano-beads became significantly wider, as reported in Fig. 4e. By using a non-uniform DPM (see Fig. 4b), which represents this sample tilt, *Deep-Z* can act on the blurred input image (Fig. 4c) and accurately bring all the nano-beads into focus (Fig. 4d), *even though it was only trained using uniform DPMs*. The lateral FWHM values calculated at the network output image became monodispersed, with a median of ~ 0.96 μm (Fig. 4f), in comparison to a median of ~ 2.14 μm at the input image. Similarly, Fig. 4g illustrates the second measurement, where the nano-beads were distributed on a *cylindrical* surface with a diameter of ~7.2 mm. As a result, the measured raw fluorescence image exhibited defocused regions as illustrated in Fig. 4i, and the FWHM values of these nano-bead images were accordingly broadened (Fig. 4k), corresponding to a median value of ~ 2.41 μm. On the other hand, using a non-uniform DPM that defines this cylindrical surface (Fig. 4h), the aberration in Fig. 4i was corrected using *Deep-Z* (Fig. 4j), and similar to the tilted sample case, the lateral FWHM values calculated at the network output image once again became monodispersed, as desired, with a median of ~ 0.91 μm (Fig. 4l).

To evaluate the limitations of this technique, we quantified the 3D surface curvature that a DPM can have without generating artifacts. For this, we used a series of DPMs that consisted of 3D sinusoidal patterns with lateral periods of $D = 1, 2, …, 256$ pixels along the x-direction (with a pixel size of 0.325 μm) and an axial oscillation range of 8 μm, i.e., a sinusoidal depth span of -1 μm to -9 μm with respect to the input plane. Each one of these 3D sinusoidal DPMs was appended on an input fluorescence image that was fed into the *Deep-Z* network. The network output at each sinusoidal 3D surface defined by the corresponding DPM was then compared against the images that were interpolated in 3D using an axially-scanned z-stack with a scanning step size of 0.5 µm, which formed the ground truth images that we used for comparison. As summarized in Supplementary Fig. S7, the *Deep-Z* network can reliably refocus the input fluorescence image onto 3D surfaces defined by sinusoidal DPMs when the period of the modulation is > 100 pixels (i.e., > 32 µm in object space). For faster oscillating DPMs, with periods smaller than 32 µm, the network output images at the corresponding 3D surfaces exhibit background modulation at these high-frequencies and their harmonics as illustrated in the spectrum analysis reported in Supplementary Fig. S7. These higher harmonic artifacts and the background modulation disappear for lower frequency DPMs, which define sinusoidal 3D surfaces at the output with a lateral period of > 32 µm and an axial range of 8 µm.

**Cross-modality digital refocusing of fluorescence images: *Deep-Z+***

*Deep-Z* framework enables digital refocusing of out-of-focus 3D features in a wide-field fluorescence microscope image to user-defined surfaces. The same concept can also be used to perform *cross-modality* digital refocusing of an input fluorescence image, where the generator network can be trained using pairs of input and label images captured by *two different* fluorescence imaging modalities, which we term as *Deep-Z+*. After its training, the *Deep-Z+* network learns to digitally refocus a single input fluorescence image acquired by a fluorescence microscope to a user-defined target surface in 3D, but this time the output will match an image of the same sample captured by a *different* fluorescence imaging modality at the corresponding height/plane. To demonstrate this unique capability, we trained a *Deep-Z+* network using pairs of wide-field microscopy images (used as inputs) and confocal microscopy images at the corresponding planes (used as ground truth labels) to perform cross-modality digital refocusing (see the Methods for training details). Fig. 5 demonstrates our blind testing results for imaging microtubule structures of BPAEC using this *Deep-Z+* framework. As seen in Fig. 5, the trained *Deep-Z+* network digitally refocused the input wide field fluorescence image onto different axial distances, while at the same time rejecting some of the defocused spatial features at the refocused planes, matching the confocal images of the corresponding planes, which serve as our ground truth. For instance, the microtubule structure at the lower left corner of a ROI in Fig. 5, which was prominent at a refocusing distance of z = 0.34 µm, was digitally rejected by *Deep-Z+* at a refocusing distance of $z$ = -0.46 µm since it became out-of-focus at this axial distance, matching the corresponding image of the confocal microscope at the same depth. As demonstrated in Fig. 5, *Deep-Z+* merges the sectioning capability of confocal microscopy with its image refocusing framework. Fig. 5 also reports *x-z* and *y-z* cross-sections of the *Deep-Z+* output images, where the axial distributions of the microtubule structures are significantly sharper in comparison to the axial scanning images of a wide-field fluorescence microscope, providing a very good match to the cross-sections obtained with a confocal microscope, matching the aim of its training.

**Discussion**

We developed a unique framework (*Deep-Z*), powered by deep neural networks, that enables 3D refocusing within a sample using a single 2D fluorescence image. This framework is non-iterative and does not require hyperparameter tuning following its training stage. In *Deep-Z*, the user can specify refocusing distances for each pixel in a DPM (following the axial range used in the training), and the fluorescence image can be digitally refocused to the corresponding surface through *Deep-Z*, within the transformation limits reported earlier in our Results section (see e.g., Supplementary Figs. S2 and S7). As illustrated in the Supplementary Information, the *Deep-Z* framework is also robust to changes in the density of the fluorescent objects within the sample volume (up to a limit, which is a function of the axial refocusing distance), the exposure time of the input images, as well as the illumination intensity modulation (see Supplementary Figs. S10,13,15-17 for detailed analyses). Because the distances are encoded in DPM and modeled as a convolutional channel, we can train the network with uniform DPMs, which still permits us to apply various non-uniform DPMs during the inference stage as reported in the Results section for e.g., correcting the sample drift, tilt, curvature or other optical aberrations, which brings additional degrees-of-freedom to the imaging system.

Deep learning has also been recently demonstrated to be very effective in performing deconvolution to boost the lateral [32–37] and the axial [38,39] resolution in microscopy images. *Deep-Z* network reported here is unique as it *selectively* deconvolves the spatial features that come into focus through the digital refocusing process (see e.g. Supplementary Fig. S5), while convolving other features that go out-of-focus, bringing the contrast to in-focus features, based on a user-defined DPM. Through this *Deep-Z* framework, we bring the snapshot *3D* refocusing capability of coherent imaging and holography to incoherent fluorescence microscopy, without any mechanical scanning, additional hardware components, or a trade-off of imaging resolution or speed. This not only significantly boosts the imaging speed, but also reduces the negative effects of photobleaching and phototoxicity on the sample. For a widefield fluorescence microscopy experiment, where an axial image stack is acquired, the illumination excites the fluorophores through the entire thickness of the specimen, and the total light exposure of a given point within the sample volume is proportional to the number of imaging planes ($N_z$) that are acquired during a single-pass z-stack. In contrast, *Deep-Z* only requires a single image acquisition step, if its axial training range covers the sample depth. Therefore, this reduction, enabled by *Deep-Z*, in the number of axial planes that need to be imaged within a sample volume directly helps to reduce the photodamage to the sample (refer to Supplementary Fig. S18 and the Supplementary Information, the sub-section titled "*Reduced photodamage using Deep-Z*" for details).

Finally, we should note that the retrievable axial range in this method depends on the SNR of the recorded image, i.e., if the depth information carried by the PSF falls below the noise floor, accurate inference will become a challenging task. To validate the performance of a pre-trained *Deep-Z* network model under variable SNR, we tested the inference of *Deep-Z* under different exposure conditions (Supplementary Fig. S10), revealing the robustness of its inference over a broad range of image exposure times that were not included in the training data (refer to the Supplementary Information, sub-section "*Deep-Z virtual refocusing capability at lower image exposure*" for details). In our Results section, we demonstrated an enhancement of ~20× in the DOF of a wide-field fluorescence image using *Deep-Z*. This axial refocusing range is in fact not an absolute limit but rather a practical choice for our training data, and it may be further improved through hardware modifications to the optical set-up by e.g., engineering the PSF in the axial direction [7,15,40–42] (see Supplementary Video S10 for an experimental demonstration of *Deep-Z* blind inference for a double-helix PSF). In addition to requiring extra hardware and sensitive alignment/calibration, such approaches would also require brighter fluorophores, to compensate for photon losses due to the insertion of additional optical components in the detection path.

## Methods

### Sample preparation

The 300 nm red fluorescence nano-beads were purchased from MagSphere Inc. (Item # PSF-300NM 0.3 UM RED), diluted by 5,000 times with methanol, and ultrasonicated for 15 minutes before and after dilution to break down the clusters. For the fluorescent bead samples on a flat surface and a tilted surface, a #1 coverslip (22×22 mm$^2$, ~150 μm thickness) was thoroughly cleaned and plasma treated. Then, a 2.5 μL droplet of the diluted bead sample was pipetted onto the coverslip and dried. For the fluorescent bead sample on a curved (cylindrical) surface, a glass tube (~ 7.2 mm diameter) was thoroughly cleaned and plasma treated. Then a 2.5 μL droplet of the diluted bead sample was pipetted onto the outer surface of the glass tube and dried.

Structural imaging of *C. elegans* neurons was carried out in strain AML18. AML18 carries the genotype wtfIs3 [rab-3p::NLS::GFP + rab-3p::NLS::tagRFP] and expresses GFP and tagRFP in the nuclei of all the neurons [29]. For functional imaging, we used the strain AML32, carrying wtfIs5 [rab-3p::NLS::GCaMP6s + rab-3p::NLS::tagRFP]. The strains were acquired from the Caenorhabditis Genetics Center (CGC). Worms were cultured on Nematode Growth Media (NGM) seeded with OP50 bacteria using standard conditions [43]. For imaging, worms were washed off the plates with M9, and anaesthetized with 3 mM levamisole [44]. Anaesthetized worms were then mounted on slides seeded with 3% Agarose. To image moving worms, the levamisole was omitted.

Two slides of multi-labeled bovine pulmonary artery endothelial cells (BPAEC) were acquired from Thermo Fisher: FluoCells Prepared Slide #1 and FluoCells Prepared Slide #2. These cells were labeled to express different cell structures and organelles. The first slide uses Texas Red for mitochondria and FITC for F-actin structures. The second slide uses FITC for microtubules.

**Fluorescence image acquisition**

The fluorescence images of nano-beads, *C. elegans* structure and BPAEC samples were captured by an inverted scanning microscope (IX83, Olympus Life Science) using a 20×/0.75NA objective lens (UPLSAPO20X, Olympus Life Science). A 130W fluorescence light source (U-HGLGPS, Olympus Life Science) was used at 100% output power. Two bandpass optical filter sets were used: Texas Red and FITC. The bead samples were captured by placing the coverslip with beads directly on the microscope sample mount. The tilted surface sample was captured by placing the coverslip with beads on a 3D-printed holder, which creates a 1.5° tilt with respect to the focal plane. The cylindrical tube surface with fluorescent beads was placed directly on the microscope sample mount. These fluorescent bead samples were imaged using Texas Red filter set. The *C. elegans* sample slide was placed on the microscope sample mount and imaged using Texas Red filter set. The BPAEC slide was placed on the microscope sample mount and imaged using Texas Red and FITC filter sets. For all the samples, the scanning microscope had a motorized stage (PROSCAN XY STAGE KIT FOR IX73/83) that moved the samples to different FOVs and performed image-contrast-based auto-focus at each location. The motorized stage was controlled using MetaMorph® microscope automation software (Molecular Devices, LLC). At each location, the control software autofocused the sample based on the standard deviation of the image, and a z-stack was taken from -20 µm to 20 µm with a step size of 0.5 µm. The image stack was captured by a monochrome scientific CMOS camera (ORCA-flash4.0 v2, Hamamatsu Photonics K.K), and saved in non-compressed tiff format, with 81 planes and 2048 × 2048 pixels in each plane.

The images of *C. elegans* neuron activities were captured by another scanning wide-field fluorescence microscope (TCS SP8, Leica Microsystems) using a 20×/0.8NA objective lens (HCPLAPO20x/0.80DRY, Leica Microsystems) and a 40×/1.3NA objective lens (HC PL APO 40x/1.30 OIL, Leica Microsystems). Two bandpass optical filter sets were used: Texas Red and FITC. The images were captured by a monochrome scientific CMOS camera (Leica-DFC9000GTC-VSC08298). For capturing image stacks of anesthetized worms, the motorized stage controlled by a control software (LAS X, Leica Microsystems) moved the sample slide to different FOVs. At each FOV, the control software took a z-stack from -20 µm to 20 µm with a step size of 0.5 µm for the 20×/0.8NA objective lens images, and with a step size of 0.27 µm for the 40×/1.3NA objective lens images, with respect to a middle plane (z=0 µm). Two images were taken at each z-plane, for Texas Red channel and FITC channel respectively. For capturing 2D videos of dynamic worms, the control software took a time-lapsed video that also time-multiplexed the Texas Red and FITC channels at the maximum speed of the system. This resulted in an average framerate of ~3.6 fps for a maximum camera framerate of 10 fps, for imaging both channels.

The BPAEC wide-field and confocal fluorescence images were captured by another inverted scanning microscope (TCS SP5, Leica Microsystems). The images were acquired using a 63×/1.4 NA objective lens (HC PL APO 63x/1.40 Oil CS2, Leica Microsystems) and FITC filter set was used. The wide-field images were recorded by a CCD with 1380×1040 pixels and 12-bit dynamic range, whereas the confocal images were recorded by a photo-multiplier tube (PMT) with 8-bit dynamic range (1024×1024 pixels). The scanning microscope had a motorized stage that moved the sample to different FOVs and depths. For each location, a stack of 12 images with 0.2 µm axial spacing was recorded.

**Image pre-processing and training data preparation**

Each captured image stack was first axially aligned using an ImageJ plugin named "StackReg" [45], which corrects the rigid shift and rotation caused by the microscope stage inaccuracy. Then an extended depth of field (EDF) image was generated using another ImageJ plugin named "Extended Depth of Field" [46]. This EDF image was used as a reference image to normalize the whole image stack, following three steps: (1) Triangular threshold[47] was used on the image to separate the background and foreground pixels; (2) the mean intensity of the background pixels of the EDF image was determined to be the background noise and subtracted; (3) the EDF image intensity was scaled to 0-1, where the scale factor was determined such that 1% of the foreground pixels above the background were greater than one (i.e., saturated); and (4) each image in the stack was subtracted by this background level and normalized by this intensity scaling factor. For testing data without an image stack, steps (1) – (3) were applied on the input image instead of the EDF image.

To prepare the training and validation datasets, on each FOV, a geodesic dilation [48] with fixed thresholds was applied on fluorescence EDF images to generate a mask that represents the regions containing the sample fluorescence signal above the background. Then, a customized greedy algorithm was used to determine a minimal set of regions with $256 \times 256$ pixels that covered this mask, with ~5% area overlaps between these training regions. The lateral locations of these regions were used to crop images on each height of the image stack, where the middle plane for each region was set to be the one with the highest standard deviation. Then 20 planes above and 20 planes below this middle plane were set to be the range of the stack, and an input image plane was generated from each one of these 41 planes. Depending on the size of the data set, around 5-10 out of these 41 planes were randomly selected as the corresponding target plane, forming around 150 to 300 image pairs. For each one of these image pairs, the refocusing distance was determined based on the location of the plane (i.e. 0.5 µm times the difference from the input plane to the target plane). By repeating this number, a uniform DPM was generated and appended to the input fluorescence image. The final dataset typically contained ~ 100,000 image pairs. This was randomly divided into a training dataset and a validation dataset, which took 85% and 15% of the data respectively. During the training process, each data point was further augmented five times by flipping or rotating the images by a random multiple of 90°. The validation dataset was not augmented. The testing dataset was cropped from separate measurements with sample FOVs that do not overlap with the FOVs of the training and validation data sets.

### *Deep-Z* network architecture

The *Deep-Z* network is formed by a least square GAN (LS-GAN) framework [49], and it is composed of two parts: a generator and a discriminator, as shown in Supplementary Fig. S8. The generator is a convolutional neural network (CNN) inspired by the U-Net [50], and follows a similar structure as in Ref. [51]. The generator network consists of a down-sampling path and a symmetric up-sampling path. In the down sampling path, there are five down-sampling blocks. Each block contains two convolutional layers that map the input tensor $x_k$ to the output tensor $x_{k+1}$:

$$x_{k+1} = x_k + \text{ReLU}[\text{CONV}_{k_2}\{\text{ReLU}[\text{CONV}_{k_1}\{x_k\}]\}] \quad (1)$$

where ReLU[.] stands for the rectified linear unit operation, and CONV{.} stands for the convolution operator (including the bias terms). The subscript of CONV denotes the number of channels in the convolutional layer; along the down-sampling path we have: $k_1 = 25, 72, 144, 288, 576$ and $k_2 = 48, 96, 192, 384, 768$ for levels $k = 1, 2, 3, 4, 5$, respectively. The "+" sign in Eq. (S1) represents a residual connection. Zero padding was used on the input tensor $x_k$ to compensate for the channel number mismatch between the input and output tensors. The connection between two consecutive down-sampling blocks is a 2×2 max-pooling layer with a stride of 2×2 pixels to perform a 2× down-sampling. The fifth down-sampling block connects to the up-sampling path, which will be detailed next.

In the up-sampling path, there are four corresponding up-sampling blocks, each of which contains two convolutional layers that map the input tensor $y_{k+1}$ to the output tensor $y_k$ using:

$$y_k = \text{ReLU}[\text{CONV}_{k_4}\{\text{ReLU}[\text{CONV}_{k_3}\{\text{CAT}(x_{k+1}, y_{k+1})\}]\}] \quad (2)$$

where the CAT(·) operator represents the concatenation of the tensors along the channel direction, i.e. $\text{CAT}(x_{k+1}, y_{k+1})$ appends tensor $x_{k+1}$ from the down-sampling path to the tensor $y_{k+1}$ in the up-sampling path at the corresponding level k+1. The number of channels in the convolutional layers, denoted by $k_3$ and $k_4$, are $k_3 = 72, 144, 288, 576$ and $k_4 = 48, 96, 192, 384$ along the up-sampling path for $k = 1, 2, 3, 4$, respectively. The connection between consecutive up-sampling blocks is an up-convolution (convolution transpose) block that up-samples the image pixels by 2×. The last block is a convolutional layer that maps the 48 channels to one output channel (see Supplementary Fig. S8).

The discriminator is a convolutional neural network that consists of six consecutive convolutional blocks, each of which maps the input tensor $z_i$ to the output tensor $z_{i+1}$, for a given level i:

$$z_{i+1} = \text{LReLU}[\text{CONV}_{i_2}\{\text{LReLU}[\text{CONV}_{i_1}\{z_i\}]\}] \quad (3)$$

where the LReLU stands for leaky ReLU operator with a slope of 0.01. The subscript of the convolutional operator represents its number of channels, which are $i_1 = 48, 96, 192, 384, 768, 1536$ and $i_2 = 96, 192, 384, 768, 1536, 3072$, for the convolution block $i = 1, 2, 3, 4, 5, 6$, respectively.

After the last convolutional block, an average pooling layer flattens the output and reduces the number of parameters to 3072. Subsequently there are fully-connected (FC) layers of size $3072 \times 3072$ with LReLU activation functions, and another FC layer of size $3072 \times 1$ with a Sigmoid activation function. The final output represents the discriminator score, which falls within (0, 1), where 0 represents a false and 1 represents a true label.

All the convolutional blocks use a convolutional kernel size of $3 \times 3$ pixels, and replicate padding of one pixel unless mentioned otherwise. All the convolutions have a stride of $1 \times 1$ pixel, except the second convolutions in Eq. (S3), which has a stride of $2 \times 2$ pixels to perform a 2× down-sampling in the discriminator path. The weights are initialized using the *Xavier* initializer [52], and the biases are initialized to 0.1.

**Training and testing of the *Deep-Z* network**

The *Deep-Z* network learns to use the information given by the appended DPM to digitally refocus the input image to a user-defined plane. In the training phase, the input data of the generator $G(.)$ have the dimensions of $256 \times 256 \times 2$, where the first channel is the fluorescence image, and the second channel is the user-defined DPM. The target data of $G(.)$ have the dimensions of $256 \times 256$, which represent the corresponding fluorescence image at a surface specified by the DPM. The input data of the discriminator $D(.)$ have the dimensions of $256 \times 256$, which can be either the generator output or the corresponding target $z^{(i)}$. During the training phase, the network iteratively minimizes the generator loss $L_G$ and discriminator loss $L_D$, defined as:

$$L_G = \frac{1}{2N} \cdot \sum_{i=1}^{N} \left[D\left(G(x^{(i)})\right) - 1\right]^2 + \alpha \cdot \frac{1}{2N} \cdot \sum_{i=1}^{N} MAE(x^{(i)}, z^{(i)}) \quad (4)$$

$$L_D = \frac{1}{2N} \cdot \sum_{i=1}^{N} \left[D\left(G(x^{(i)})\right)\right]^2 + \frac{1}{2N} \cdot \sum_{i=1}^{N} \left[D(z^{(i)}) - 1\right]^2 \quad (5)$$

where $N$ is the number of images used in each batch (e.g., $N = 20$), $G(x^{(i)})$ is the generator output for the input $x^{(i)}$, $z^{(i)}$ is the corresponding target label, $D(.)$ is the discriminator, and MAE(.) stands for mean absolute error. $\alpha$ is a regularization parameter for the GAN loss and the MAE loss in $L_G$. In the training phase, it was chosen as $\alpha = 0.02$. For training stability and optimal performance, adaptive momentum optimizer (Adam) was used to minimize both $L_G$ and $L_D$, with a learning rate of $10^{-4}$ and $3 \times 10^{-5}$ for $L_G$ and $L_D$ respectively. In each iteration, six updates of the generator loss and three updates of the discriminator loss were performed. The validation set was tested every 50 iterations, and the best network (to be blindly tested) was chosen to be the one with the smallest MAE loss on the validation set.

In the testing phase, once the training is complete, only the generator network is active. Limited by the graphical memory of our GPU, the largest image FOV that we tested was $1536 \times 1536$ pixels. Because image was normalized to be in the range 0–1, whereas the refocusing distance was on the scale of around -10 to 10 (in units of µm), the DPM entries were divided by 10 to be in the range of -1 to 1 before the training and testing of the *Deep-Z* network, to keep the dynamic range of the image and DPM matrices similar to each other.

The network was implemented using TensorFlow [53], performed on a PC with Intel Core i7-8700K six-core 3.7GHz CPU and 32GB RAM, using an Nvidia GeForce 1080Ti GPU. On average, the training takes ~ 70 hours for ~ 400,000 iterations (equivalent to ~ 50 epochs). After the training, the network inference time was ~ 0.2 s for an image with 512 × 512 pixels and ~ 1s for an image with 1536 × 1536 pixels on the same PC.

**Measurement of the lateral and axial FWHM values of the fluorescent beads samples.**

For characterizing the lateral FWHM of the fluorescent beads samples, a threshold was performed on the image to extract the connected components. Then, individual regions of 30 × 30 pixels were cropped around the centroid of these connected components. A 2D Gaussian fit was performed on each of these individual regions, which was done using lsqcurvefit [54] in Matlab (MathWorks, Inc) to match the function:

$$I(x,y) = A \cdot \exp\left[\frac{(x-x_c)^2}{2 \cdot \sigma_x^2} + \frac{(y-y_c)^2}{2 \cdot \sigma_y^2}\right] \quad (6)$$

The lateral FWHM was then calculated as the mean FWHM of x and y directions, i.e.,

$$\text{FWHM}_{\text{lateral}} = 2\sqrt{2\ln 2} \cdot \frac{\sigma_x \cdot \Delta_x + \sigma_y \cdot \Delta_y}{2} \quad (7)$$

where $\Delta_x = \Delta_y = 0.325 \; \mu m$ was the effective pixel size of the fluorescence image on the object plane. A histogram was subsequently generated for the lateral FWHM values for all the thresholded beads (e.g., n = 461 for Fig. 1 and n > 750 for Fig. 4).

To characterize the axial FWHM values for the bead samples, slices along the x-z direction with 81 steps were cropped at $y = y_c$ for each bead, from either the digitally refocused or the mechanically-scanned axial image stack. Another 2D Gaussian fit was performed on each cropped slice, to match the function:

$$I(x, z) = A \cdot \exp\left[\frac{(x - x_c)^2}{2 \cdot \sigma_x^2} + \frac{(z - z_c)^2}{2 \cdot \sigma_z^2}\right] \quad (8)$$

The axial FWHM was then calculated as:

$$\text{FWHM}_{\text{axial}} = 2\sqrt{2 \ln 2} \cdot \sigma_z \cdot \Delta_z \quad (9)$$

where $\Delta_z = 0.5 \; \mu m$ was the axial step size. A histogram was subsequently generated for the axial FWHM values.

**Image quality evaluation**

The network output images $I^{\text{out}}$ were evaluated with reference to the corresponding ground truth images $I^{\text{GT}}$ using five different criteria: (1) mean square error (MSE), (2) root mean square error (RMSE), (3) MAE, (4) correlation coefficient, and (5) SSIM [55]. The MSE is one of the most widely used error metrics, defined as:

$$\text{MSE}(I^{\text{out}}, I^{\text{GT}}) = \frac{1}{N_x \cdot N_y} \left\| I^{\text{out}} - I^{\text{GT}} \right\|_2^2 \quad (10)$$

where $N_x$ and $N_y$ represent the number of pixels in the x and y directions, respectively. The square root of MSE results in RMSE. Compared to MSE, MAE uses 1-norm difference (absolute difference) instead of 2-norm difference, which is less sensitive to significant outlier pixels:

$$\text{MAE}(I^{out}, I^{GT}) = \frac{1}{N_x \cdot N_y} \left\| I^{out} - I^{GT} \right\|_1 \quad (11)$$

The correlation coefficient is defined as:

$$\text{corr}(I^{out}, I^{GT}) = \frac{\sum_x \sum_y (I_{xy}^{out} - \mu_{out})(I_{xy}^{GT} - \mu_{GT})}{\sqrt{\left(\sum_x \sum_y (I_{xy}^{out} - \mu_{out})^2\right)\left(\sum_x \sum_y (I_{xy}^{GT} - \mu_{GT})^2\right)}} \quad (12)$$

where $\mu_{out}$ and $\mu_{GT}$ are the mean values of the images $I^{out}$ and $I^{GT}$ respectively.

While these criteria listed above can be used to quantify errors in the network output compared to the GT, they are not strong indicators of the *perceived similarity* between two images. SSIM aims to address this shortcoming by evaluating the structural similarity in the images, defined as:

$$\text{SSIM}(I^{out}, I^{GT}) = \frac{(2\mu_{out}\mu_{GT} + C_1)(2\sigma_{out,GT} + C_2)}{(\mu_{out}^2 + \mu_{GT}^2 + C_1)(\sigma_{out}^2 + \sigma_{GT}^2 + C_2)} \quad (13)$$

where $\sigma_{out}$ and $\sigma_{GT}$ are the standard deviations of $I^{out}$ and $I^{GT}$ respectively, and $\sigma_{out,GT}$ is the cross-variance between the two images.

**Tracking and quantification of *C. elegans* neuron activity**

The *C. elegans* neuron activity tracking video was captured by time-multiplexing the two fluorescence channels (FITC, followed by TexasRed, and then FITC and so on). The adjacent frames were combined so that the green color channel was FITC (neuron activity) and the red color channel was Texas Red (neuron nuclei). Subsequent frames were aligned using a feature-based registration toolbox with projective transformation in Matlab (MathWorks, Inc.) to correct for slight body motion of the worms. Each input video frame was appended with DPMs representing propagation distances from -10 µm to 10 µm with 0.5 µm step size, and then tested through a *Deep-Z* network (specifically trained for this imaging system), which generated a virtual axial image stack for each frame in the video.

To localize individual neurons, the red channel stacks (Texas Red, neuron nuclei) were projected by median-intensity through the time sequence. Local maxima in this projected median intensity stack marked the centroid of each neuron and the voxels of each neuron was segmented from these centroids by watershed segmentation[48], which generated a 3D spatial voxel mask for each neuron. A total of 155 neurons were isolated. Then, the average of the 100 brightest voxels in the green channel (FITC, neuron activity) inside each neuron spatial mask was calculated as the calcium activity intensity $F_i(t)$, for each time frame $t$ and each neuron $i = 1,2,\ldots,155$. The differential activity was then calculated, $\Delta F(t) = F(t) - F_0$, for each neuron, where $F_0$ is the time average of $F(t)$.

By thresholding on the standard deviation of each $\Delta F(t)$, we selected the 70 most active cells and performed further clustering on them based on their calcium activity pattern similarity (Supplementary Fig. S6b) using a spectral clustering algorithm [56,57]. The calcium activity pattern similarity was defined as

$$S_{ij} = \exp\left(-\frac{\left\|\frac{\Delta F_i(t)}{F_{i0}} - \frac{\Delta F_j(t)}{F_{j0}}\right\|^2}{\sigma^2}\right) \quad (14)$$

for neurons i and j, which results in a similarity matrix S (Supplementary Fig. S6c). $\sigma = 1.5$ is the standard deviation of this Gaussian similarity function, which controls the width of the neighbors in the similarity graph. The spectral clustering solves an eigen-value problem on the graph Laplacian L generated from the similarity matrix S, defined as the difference of weight matrix W and degree matrix D, i.e.,

$$L = D - W \quad (15)$$

where

$$W_{ij} = \begin{cases} S_{ij} & \text{if } i \neq j \\ 0 & \text{if } i = j \end{cases} \quad (16)$$

$$D_{ij} = \begin{cases} \sum_j W_{ij} & \text{if } i = j \\ 0 & \text{if } i \neq j \end{cases} \quad (17)$$

The number of clusters was chosen using eigen-gap heuristics[57], which was the index of the largest general eigenvalue (by solving general eigen value problem $Lv = \lambda Dv$) before the eigen-gap, where the eigenvalues jump up significantly, which was determined to be k=3 (see Supplementary Fig. S6d). Then the corresponding first k=3 eigen-vectors were combined as a matrix, whose rows were clustered using standard k-means clustering[57], which resulted in the three clusters of the calcium activity patterns shown in Supplementary Fig. S6e and the rearranged similarity matrix shown in Supplementary Fig. S6f.

**Cross-modality alignment of wide-field and confocal fluorescence images**

Each stack of the wide-field/confocal pair was first self-aligned and normalized using the method described in the previous sub-section. Then the individual FOVs were stitched together using

"Image Stitching" plugin of ImageJ [58]. The stitched wide-field and confocal EDF images were then co-registered using a feature-based registration with projective transformation performed in Matlab (MathWorks, Inc) [59]. Then the stitched confocal EDF images as well as the stitched stacks were warped using this estimated transformation to match their wide-field counterparts (Supplementary Fig. S9a). The non-overlapping regions of the wide-field and warped confocal images were subsequently deleted. Then the above-described greedy algorithm was used to crop non-empty regions of $256 \times 256$ pixels from the remaining stitched wide-field images and their corresponding warped confocal images. The same feature-based registration was applied on each pair of cropped regions for fine alignment. This step provides good correspondence between the wide field image and the corresponding confocal image in the lateral directions (Supplementary Fig. S9b).

Although the axial scanning step size was fixed to be 0.2 μm, the reference zero-point in the axial direction for the wide-field and the confocal stacks needed to be matched. To determine this reference zero-point in the axial direction, the images at each depth were compared with the EDF image of the same region using structural similarity index (SSIM) [55], providing a focus curve (Supplementary Fig. S9c). A second order polynomial fit was performed on four points in this focus curve with highest SSIM values, and the reference zero-point was determined to be the peak of the fit (Supplementary Fig. S9c). The heights of wide-field and confocal stacks were then centered by their corresponding reference zero-points in the axial direction. For each wide-field image used as input, four confocal images were randomly selected from the stack as the target, and their DPMs were calculated based on the axial difference of the centered height values of the confocal and the corresponding wide-field images.

**Code availability**

Deep learning models reported in this work used standard libraries and scripts that are publicly available in TensorFlow. Through a custom-written Fiji based plugin, we provided our trained network models (together with some sample test images) for the following objective lenses: Leica HC PL APO 20x/0.80 DRY (two different network models trained on TxRd and FITC channels), Leica HC PL APO 40x/1.30 OIL (trained on TxRd channel), Olympus UPLSAPO20X - 0.75 NA (trained on TxRd channel). We made this custom-written plugin and our models publicly available through the following links:

http://bit.ly/deep-z-git

http://bit.ly/deep-z

**Acknowledgments:** The authors acknowledge Y. Luo, X. Tong, T. Liu, H. C. Koydemir and Z.S. Ballard of UCLA, as well as Leica Microsystems for their help with some of the experiments. The Ozcan Group at UCLA acknowledges the support of Koc Group, National Science Foundation, and the Howard Hughes Medical Institute. Y.W. also acknowledges the support of SPIE John Kiel scholarship. Some of the reported optical microscopy experiments were performed at the Advanced Light Microscopy/Spectroscopy Laboratory and the Leica Microsystems Center of Excellence at the California NanoSystems Institute at UCLA with funding support from NIH Shared Instrumentation Grant S10OD025017 and NSF Major Research Instrumentation grant CHE-0722519.

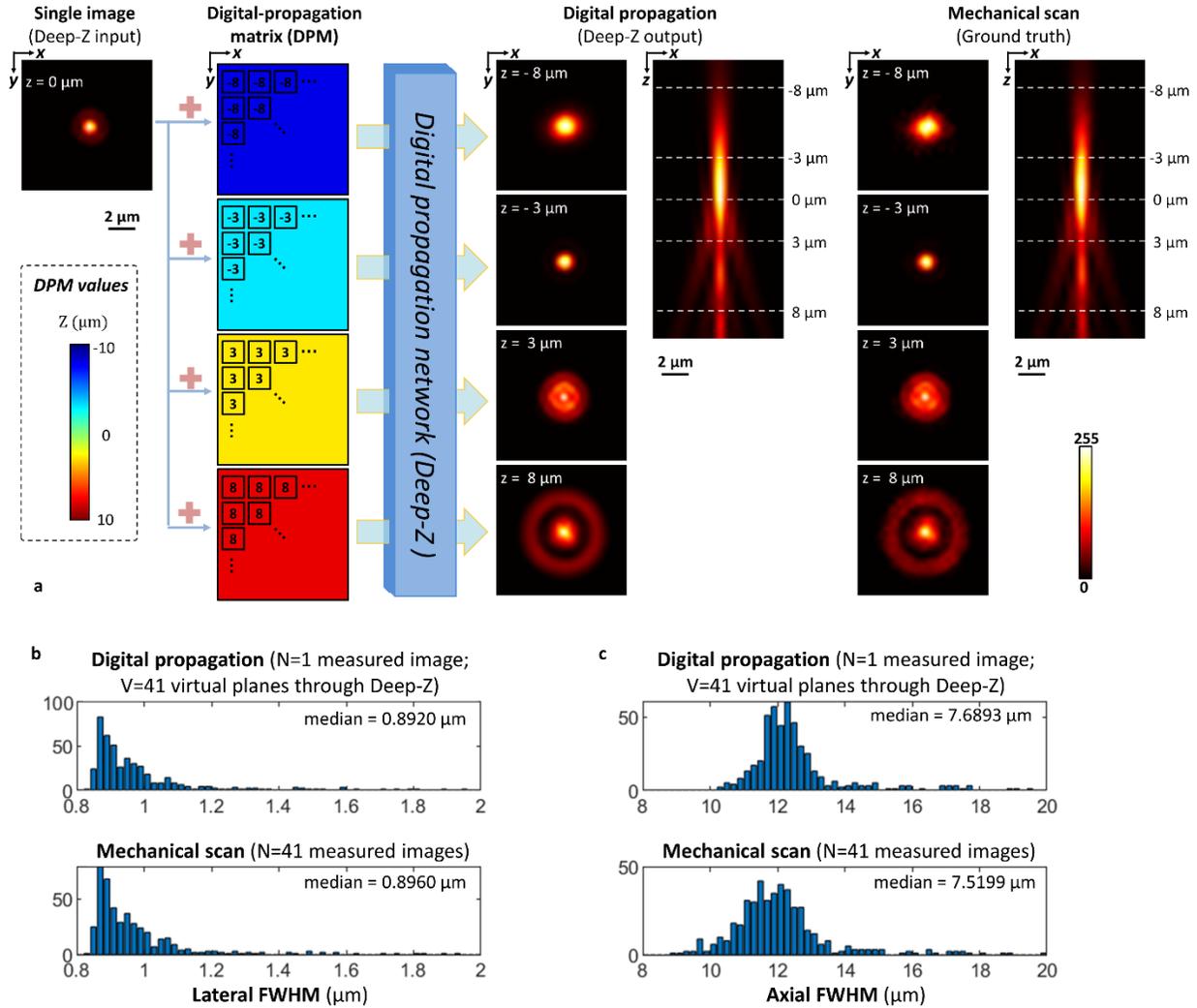

**Fig. 1. Refocusing of fluorescence images using *Deep-Z*.** (a) By concatenating a digital propagation matrix (DPM) to a single fluorescence image, and running the resulting image through a trained *Deep-Z* network, digitally refocused images at different planes can be rapidly obtained, as if an axial scan is performed at the corresponding planes within the sample volume. The DPM has the same size as the input image and its entries represent the axial propagation distance for each pixel and can also be spatially non-uniform. The results of *Deep-Z* inference are compared against the images of an axial-scanning fluorescence microscope for the same fluorescent bead (300 nm), providing a very good match. (b) Lateral FWHM histograms for 461

individual/isolated fluorescence nano-beads (300 nm) measured using *Deep-Z* inference (N=1 captured image) and the images obtained using mechanical axial scanning (N=41 captured images) provide a very good match to each other. (c) Same as in (b), except for the *axial* FWHM measurements for the same data set, also revealing a very good match between *Deep-Z* inference results and the axial mechanical scanning results.

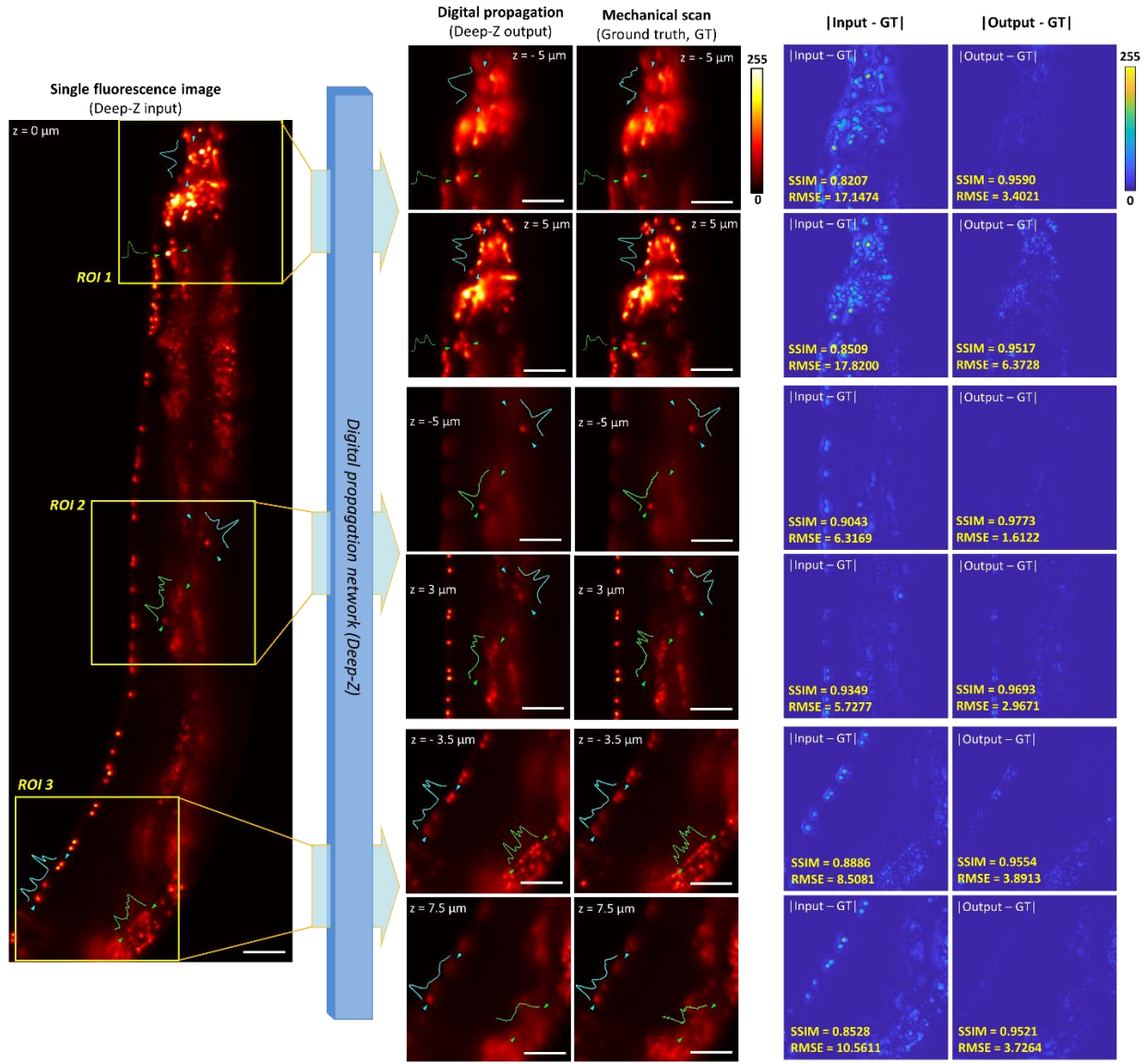

**Fig. 2. 3D imaging of *C. Elegans* neuron nuclei using *Deep-Z*.** Different ROIs are digitally refocused using *Deep-Z* to different planes within the sample volume; the resulting images provide a very good match to the corresponding ground truth images, acquired using a scanning fluorescence microscope. The absolute difference images of the input and output with respect to the corresponding ground truth image are also provided on the right, with structural similarity index (SSIM) and root mean square error (RMSE) values reported, further demonstrating the success of *Deep-Z*. Scale bar: 25 μm.

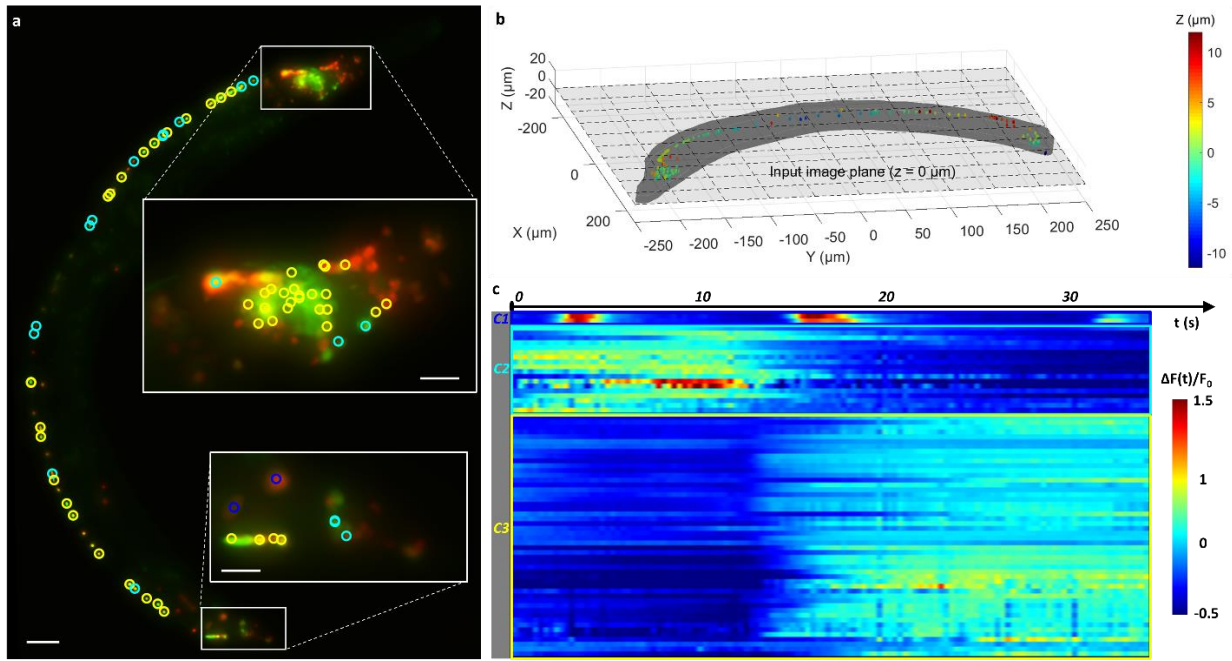

**Fig. 3. *C. Elegans* neuron activity tracking in 3D using *Deep-Z*.** (a) Maximum intensity projection (MIP) along the axial direction of the median intensity image taken across the time sequence. The red channel (Texas red) labels neuron nuclei. The green channel (FITC) labels neuron calcium activity. A total of 155 neurons were identified, 70 of which were active in calcium activity. Scale bar: 25 μm. Scale bar for the zoom-in regions: 10 μm. (b) All the 155 localized neurons are shown in 3D, where depths are color-coded. (c) 3D tracking of neuron calcium activity events corresponding to the 70 active neurons. The neurons were grouped into 3 clusters (C1-C3) based on their calcium activity pattern similarity (see Methods section). The locations of these neurons are marked by the circles in (a). The colors of the circles represent different clusters: C1(blue), C2(cyan) and C3(yellow).

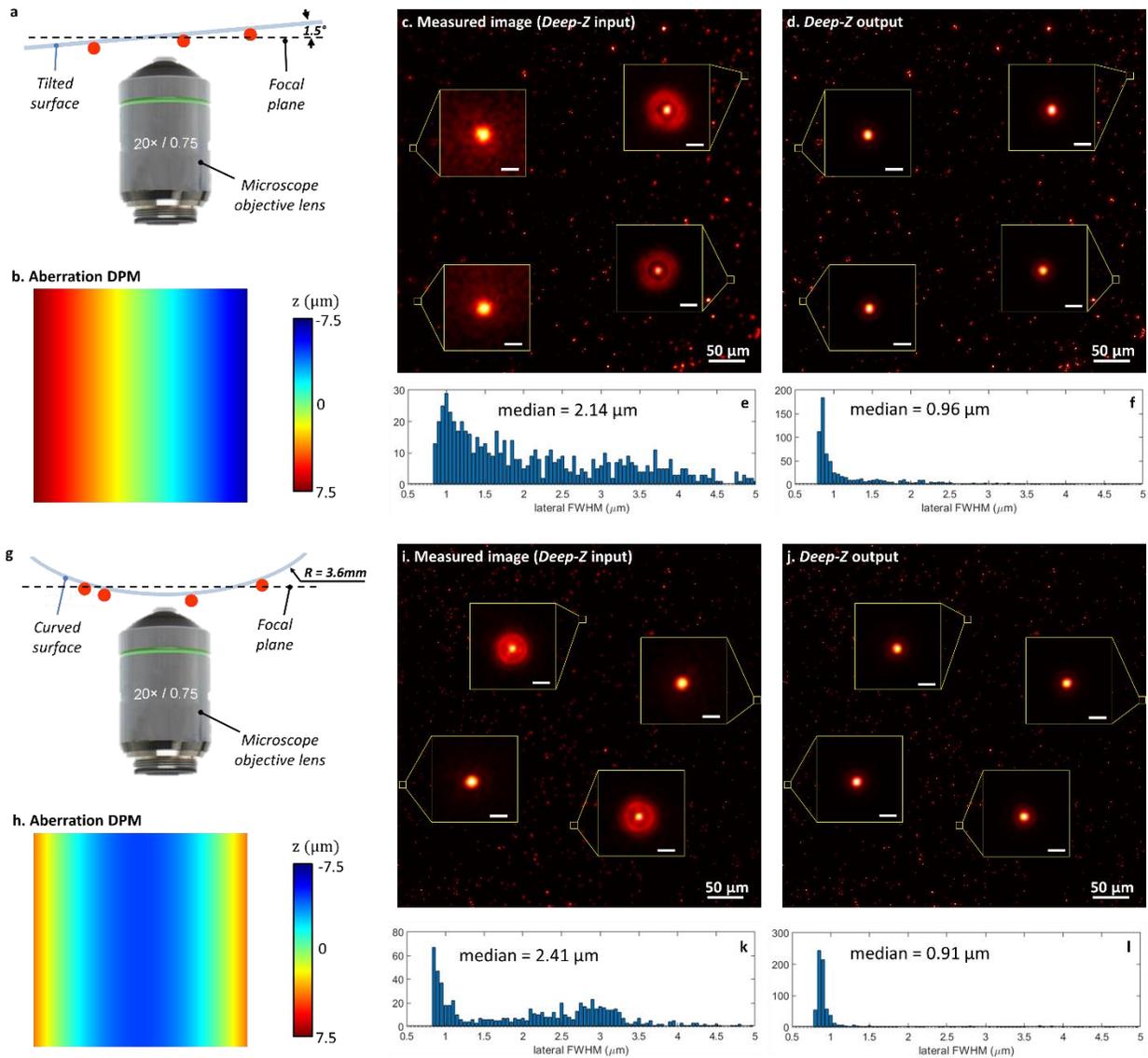

**Fig. 4. Non-uniform DPMs enable digital refocusing of a single fluorescence image onto user-defined 3D surfaces using *Deep-Z*.** (a) Measurement of a tilted fluorescent sample (300 nm beads). (b) The corresponding DPM for this tilted plane. (c) Measured raw fluorescence image; the left and right parts are out-of-focus in different directions, due to the sample tilt. (d) The *Deep-Z* output rapidly brings all the regions into correct focus. (e,f) report the lateral FWHM values of the nano-beads shown in (c,d), respectively, clearly demonstrating that *Deep-Z* with the non-uniform DPM brought the out-of-focus particles into focus. (g) Measurement of a

cylindrical surface with fluorescent beads (300 nm beads). (h) The corresponding DPM for this curved surface. (i) Measured raw fluorescence image; the middle region and the edges are out-of-focus due to the curvature of the sample. (j) The *Deep-Z* output rapidly brings all the regions into correct focus. (k,l) report the lateral FWHM values of the nano-beads shown in (i,j), respectively, clearly demonstrating that *Deep-Z* with the non-uniform DPM brought the out-of-focus particles into focus.

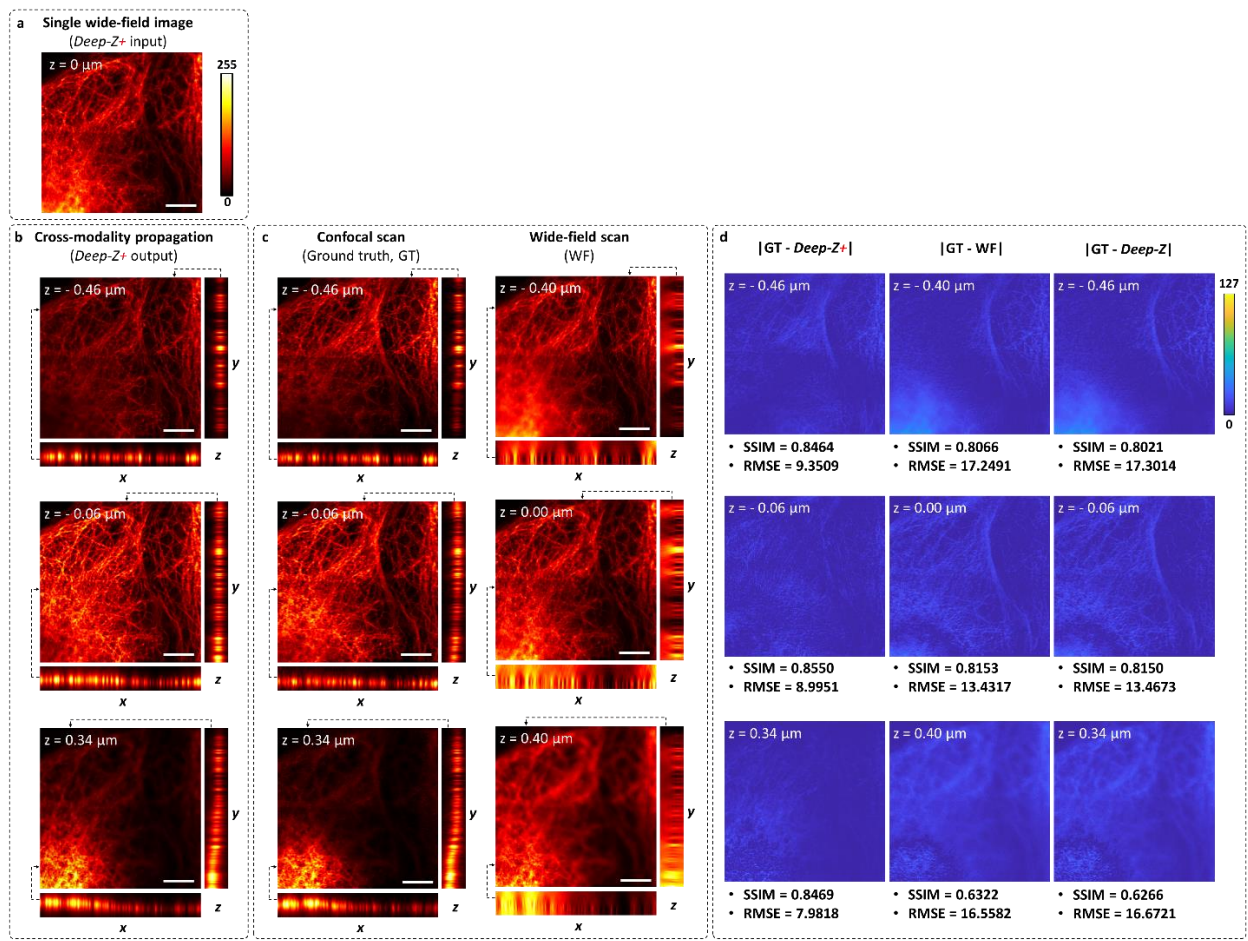

**Fig. 5. *Deep-Z+*: Cross-modality digital refocusing of fluorescence images.** A single wide-field fluorescence image (63×/1.4NA objective lens) of BPAEC microtubule structures (a) was digitally refocused using *Deep-Z+* to different planes in 3D (b), matching the images captured by a confocal microscope at the corresponding planes (c), retrieving volumetric information from a single input image and performing axial sectioning at the same time. Wide-field (WF) images are also shown in (c) for comparison. These scanning WF images report the closest heights to the corresponding confocal images, and have 60 nm axial offset since the two image stacks are discretely scanned and digitally aligned to each other. x-z and y-z cross-sections of the refocused images are also shown to demonstrate the match between *Deep-Z+* inference and the ground truth confocal microscope images of the same planes; the same cross-sections (x-z and y-z) are also shown for a wide-field scanning fluorescence microscope, reporting a significant axial blur in each case. Each cross-sectional zoomed-in image spans 1.6 µm in z-direction (with an axial step size of 0.2 µm), and the dotted arrows mark the locations, where the x-z and y-z cross-sections were taken. (d) The absolute difference images of the *Deep-Z+* output with respect to the corresponding confocal images are also provided, with SSIM and RMSE values, further quantifying the performance of *Deep-Z+*. For comparison, we also show the absolute difference images of the 'standard' *Deep-Z* output images as well as the scanning wide-field fluorescence microscope images with respect to the corresponding confocal images, both of which report increased error and weaker SSIM compared to |GT - *Deep-Z+*|. The quantitative match between |GT - WF| and |GT - *Deep-Z*| (see d) also suggests that the impact of 60 nm axial offset between the confocal and wide-field image stacks is negligible. Scale bar: 10 µm.